# BAYESIAN ONLINE CHANGE POINT DETECTION FOR BASELINE SHIFTS

**Ginga Yoshizawa**
Intel K.K.
Tokyo, 100-0005, Japan
ginga.yoshizawa@intel.com

## ABSTRACT

In time series data analysis, detecting change points on a real-time basis (online) is of great interest in many areas, such as finance, environmental monitoring, and medicine. One promising means to achieve this is the Bayesian online change point detection (BOCPD) algorithm, which has been successfully adopted in particular cases in which the time series of interest has a fixed baseline. However, we have found that the algorithm struggles when the baseline irreversibly shifts from its initial state. This is because with the original BOCPD algorithm, the sensitivity with which a change point can be detected is degraded if the data points are fluctuating at locations relatively far from the original baseline. In this paper, we not only extend the original BOCPD algorithm to be applicable to a time series whose baseline is constantly shifting toward unknown values but also visualize why the proposed extension works. To demonstrate the efficacy of the proposed algorithm compared to the original one, we examine these algorithms on two real-world data sets and six synthetic data sets.

## 1 Introduction

Online change point detection refers to the real-time detection of an acute change in a sequential time series data set. The definition of the term "change point" may vary depending on the application and the analysts's perspective. Change points can be, for instance, mean shifts [1, 8, 9, 12, 6], slope (gradient) shifts [1, 10], variance shifts [11, 6, 5, 7], anomalous events [3, 11, 4], or combinations of several of these. Regardless of the definition, however, it is reasonable to assume that the generative parameters before and after a change point are different. In actuality, the changes in the generative parameters at a change point, if substantial, should stem from changes in the parameters of the relevant physical models, which are often not trivial to articulate due to their complexity. In the present work, we concentrate on improving the performance of detecting mean shifts and slope shifts, especially when the baseline of the observed data points are continuously shifting away from the original level.

Since the first practical algorithm for Bayesian online change point detection (BOCPD) was introduced [1], it has received considerable interest for a wide range of real-world applications. These applications include water quality monitoring [3], human-machine interaction analysis [4], medical usage [5, 6], fuel management systems in unmanned vehicles [7], and satellite fault prediction [8]. Moreover, efforts have been made to improve the performance and robustness of the BOCPD algorithm itself, such as through hyper-parameter learning [9], robust change point determination [10], and the prediction of change points [11]. Since this paper aims to broaden the applications of the BOCPD, the current study is categorized as part of the latter group.

Although it has been proven that BOCPD is a promising technique for various applications, it should be noted that most previous works have dealt only with data sets whose long-term expectation is constant (e.g., a daily financial gain in %) or time series data sets with a fixed baseline (e.g., switching back and forth between normal and abnormal states). In this study, we demonstrate reasons why the original BOCPD algorithm's usage has been limited to such cases and attempt to modify the algorithm by introducing a mechanism of feeding back information on the existence of a change point to guide subsequent detection attempts.

In this paper, the term "baseline" is defined as a certain level of data values that a time series tends to regress back to ((1) and (2) in Fig.1). For instance, if we are considering an economic metric expressed as a percentage (e.g., a monthly job loss rate), it should naturally regress to a value of 0 (the baseline) over the long term. However, while the original BOCPD algorithm implicitly assumes that the values eventually return to the original baseline, this is not always true in reality if irreversible and unidirectional mean shifts occur. In other words, this paper considers cases in which parameter shifts occur in a time series, and the newly proposed extension of the BOCPD algorithm is able to detect such shifts without losing contrast.

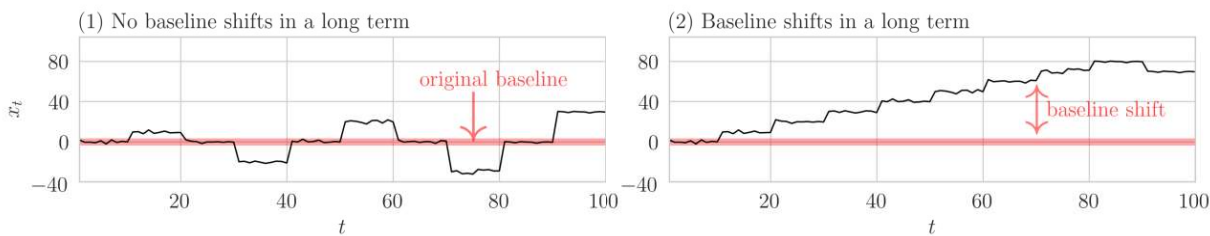

**Fig. 1. Illustrations of time series with and without baseline shifts.** If there is a substantial value shift over the long term, this is regarded as a baseline shift.

According to a comprehensive study of the categorization of change point detection algorithms [2], the keys to useful online change point detection algorithms can be summarized as follows: 1) minimal delay, 2) minimal false positives, 3) computational efficiency, and 4) robustness. This study primarily aims to improve 1), 2) and 4) without sacrificing 3), particularly in cases in which a time series data set shows unidirectional baseline shifts from a long-term perspective.

## 2 System Model

The BOCPD algorithm [1] assumes that a group of sequential data points $x_1, x_2, \ldots, x_T$, where $T \in \mathbb{N}$, in chronological order, can be partitioned into subgroups separated by estimated change points, with $\rho$ ($\in \mathbb{N}$) denoting the subgroup or partition label. To determine the interval between two change points (i.e., the length of a partition), the concept of the run length $r_t$ ($\in \mathbb{N}$) is introduced. The run length is a number that counts for how many time steps a certain partition continues without the next change point being observed. In other words, once a change point is observed, the run length $r_t$ no longer grows and is reset to zero. To determine whether the run length $r_t$ should continue to increase, the probability distribution of a certain run length, $P(r_t \mid x_{1:t})$, is calculated to estimate whether a change point has occurred whenever the next data point $x_t$ is observed. The set of subgrouped data points in partition $\rho$ is denoted by $x_t^{(\rho)}$, and the set of subgrouped data points from the beginning to a certain time $t$ is denoted by $x_{1:t}$; thus, $\bigcup_\rho x_t^{(\rho)} = x_{1:t}$. Each data point within a partition is regarded as i.i.d. and sampled from a probability distribution $P(x_t \mid \eta_\rho)$, where the parameters $\eta_\rho$ and $\rho$ are also i.i.d. In addition, the minimum $r_t$ is defined as 0 in the algorithm; therefore, the minimum number of elements that may be stored in $x_t^{(\rho)}$ is also 0.

### 2.1 BOCPD Algorithm

One of the unique mechanisms of the BOCPD algorithm is the utilization of the run-length distribution $P(r_t \mid x_{1:t})$ to assess the probability of the existence of a change point. Given a set of $t$ data points $x_{1:t}$, the run-length distribution is described as follows:

$$P(r_t \mid x_{1:t}) = \frac{P(r_t, x_{1:t})}{\sum_{r_t} P(r_t, x_{1:t})} \quad (1)$$

where the joint distribution over the run length $r_t$ and all observed data points is recursively calculated [1, 9].

$$P(r_t, x_{1:t}) = \sum_{r_{t-1}} \underbrace{P(r_t \mid r_{t-1})}_{hazard} \underbrace{P(x_t \mid r_{t-1}, x_t^{(\rho)})}_{likelihood} P(r_{t-1}, x_{1:t-1}) \quad (2)$$

For simplicity, a constant hazard function $H$ is used as a prior:

$$P(r_t \mid r_{t-1}) = \begin{cases} H(r_t + 1) & (r_t = 1) \\ 1 - H(r_{t-1} + 1) & (r_t = r_{r_t-1} + 1) \\ 0 & (otherwise) \end{cases} \quad (3)$$

$$H(r_t) = \frac{1}{\lambda} \quad (4)$$

where $\lambda$ is a parameter used to adjust the sensitivity of change point detection. This sensitivity is higher with a smaller $\lambda$ (i.e., a larger $H(r_t)$). The detailed calculation steps are described in Algorithm 1 (excluding the components marked with an asterisk (*)).

## 2.2 Exponential Family Likelihoods

As suggested in [1], exponential family likelihoods are mathematically convenient for obtaining a posterior predictive distribution $P(x_t \mid r_{t-1}, x_t^{(\rho)})$ since there exists a conjugate prior to simplify the calculation with finite parameters $\eta_\rho$. Unless a time series data set of interest is known to be based on a discrete distribution, one generic assumption that is typically used for continuous distribution is to select a normal distribution of unknown mean $\mu$ and variance $\sigma$ as a likelihood. To address this assumption, in previous studies [9, 13], a normal inverse gamma ($NIG$) distribution has been adopted as a conjugate prior. In the same way, we also select an $NIG$ distribution throughout this paper, and according to [14], the posterior predictive distribution $P(x_t \mid r_{t-1}, x_t^{(\rho)})$ can be written as shown below.

$$P(x_t \mid r_{t-1}, x_t^{(\rho)}) = t_{2\alpha}(x_t \mid \mu_t, \frac{\beta_t(\nu_t + 1)}{\nu_t \alpha_t}) \quad (5)$$

This posterior predictive distribution is a Student's t-distribution at time $t$ parameterized by mean $\mu_t$, variance $\nu_t$, $2\alpha_t$ degrees of freedom, and $\beta_t$ data points. These parameters are updated for every observation of a new data point at time $t+1$.

$$\mu_{t+1} = \frac{\nu_t \mu_t + x_t}{\nu_t + 1} \quad (6)$$

$$\nu_{t+1} = \nu_t + 1 \quad (7)$$

$$\alpha_{t+1} = \alpha_t + \frac{1}{2} \quad (8)$$

$$\beta_{t+1} = \beta_t + \frac{\nu_t (x_t - \mu_t)^2}{2(\nu_t + 1)} \quad (9)$$

As an example, the initial parameter values may be set to $\mu_1 = 0$ and $\nu_1 = \alpha_1 = \beta_1 = 1$ under the assumption that the data points in the time series start near zero and, after some fluctuations, will eventually regress back to the original baseline. In other words, this one-time setting of the initial conditions is based on the implicit hope that no irreversible, unidirectional value shifts will occur from a long-term perspective so that the model will continue to work well, as we see in later sections.

## 2.3 Change Point Detection

Regarding change point determination criteria, a variety of decision-making mechanisms have been proposed [3, 5, 6, 10, 11], and the original literature [1] did not explicitly specify any methodology for determining change points. However, in this paper, we attempt to be generic and thus simply define the existence of a change point by taking the difference $\delta$ between the indices of the highest probability among all run lengths at the current time $t$ and the one-step-previous time $t - 1$. This straightforward concept of an "argmax" criterion for determining change points has also been adopted in some previous studies [3, 5, 6].

$$\delta = argmax(P(r_t \mid x_{1:t})) - argmax(P(r_{t-1} \mid x_{1:t-1})) \quad (10)$$

If $\delta$ is positive, this indicates that the run length $r_t$ is continuing to grow, whereas either a zero or negative value indicates the existence of a change point $c_t (\in [0, 1])$.

$$c_t = \begin{cases} 0 & (\delta > 0) \\ 1 & (\delta \leq 0) \end{cases} \quad (11)$$

The flow chart in Fig. 2 describes the steps of the calculation in the BOCPD algorithm, from initializing the parameters $\eta_\rho$ and observing a new data point $x_t$ through calculating the run-length distribution $P(r_t \mid x_{1:t})$, followed by the change point decision mechanism described above.

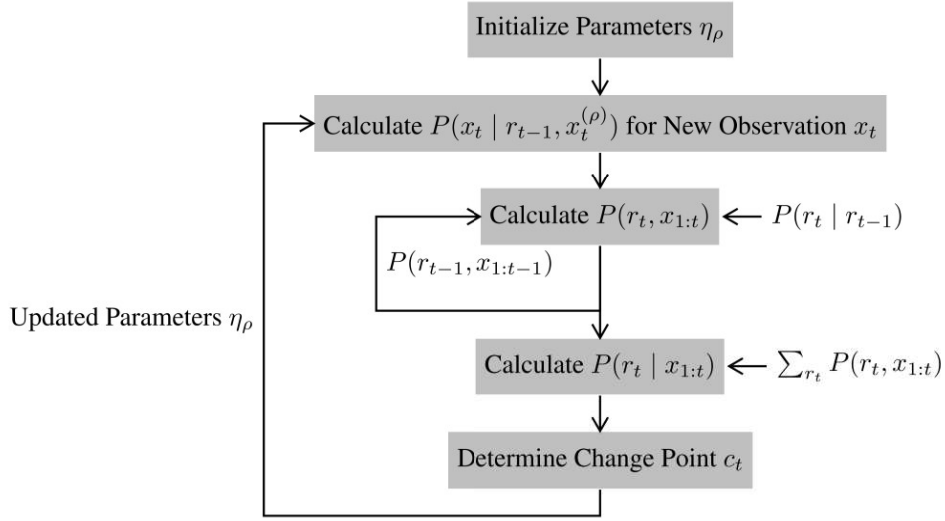

Fig. 2. The BOCPD algorithm.

### 2.4 Qualitative Change Point Detection Sensitivity Analysis

One of the critical components of the present work is the visualization of the change point detection sensitivity over time. to make the discussion of the sensitivity as intuitive as possible, we visualize the predictive probability $P(x_t \mid r_{t-1}, x_t^{(\rho)})$ accompanying data point $x_t$ to illustrate the interactions between the incoming data points and the underlying probability distribution. This is particularly useful for understanding when BOCPD starts to lose its sensitivity for change point detection. As an example, Fig. 3 illustrates how the predictive probability $P(x_t \mid r_{t-1}, x_t^{(\rho)})$ transforms with incoming data points, which eventually defines the contrast of the run-length distribution $P(r_t \mid x_{1:t})$.

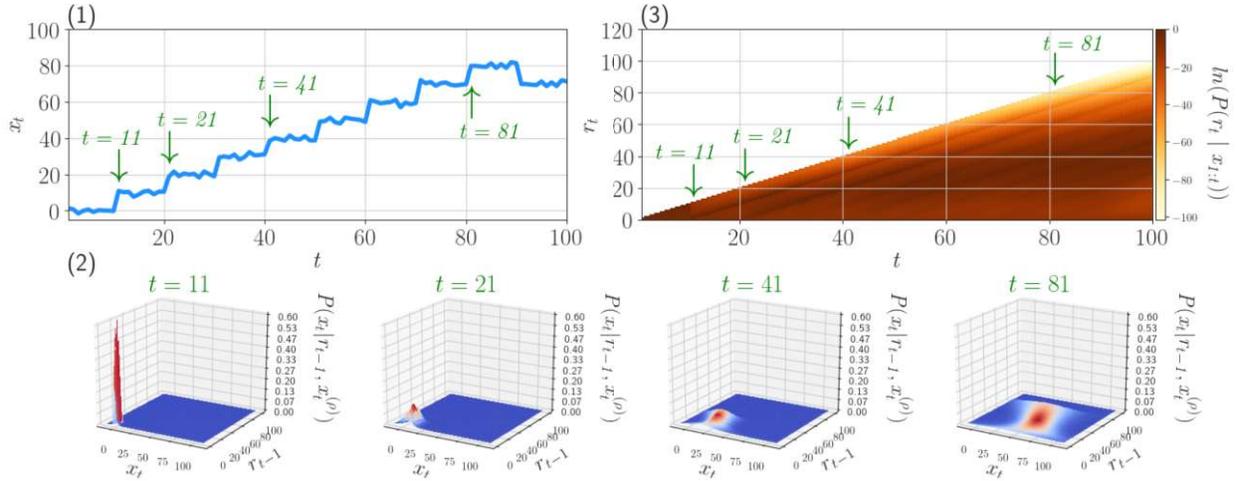

**Fig. 3. Raw data and corresponding predictive distribution and run-length distribution.** (1) A synthetic data set with predetermined change points at $t = \{11, 21, ..., 91\}$, separated by intervals of $t = 10$. The solid line indicates the data transitions and four change points are marked with arrows. (2) The predictive distribution of $x_t$ given $r_{r-1}$, mapped at $t = \{11, 21, 41, 81\}$. As the run length $r_t$ grows, with increasing distance of $x_t$ from the original baseline, the shape of the distribution becomes flatter. (3) Resulting run-length distribution over all observed data points on a natural logarithmic scale as calculated via BOCPD. Darker colors indicate higher probabilities. The dark streak becomes blurrier as time progresses, indicating a loss in the contrast needed for the determination of a change point.

## 3 Problem Statement

If there is no guarantee that the long-term expectation of a time series is approximately constant nor that an anomalous state will regress to a fixed known baseline value, the original BOCPD algorithm struggles to maintain its change point detection sensitivity. This is because as the baseline irreversibly shifts toward arbitrary values and becomes farther from the original mean, the shape of the underlying distributional assumption for determining the predictive distribution of the current data point, $P(x_t|r_{t-1}, x_t^{(\rho)})$, becomes flatter and loses its contrast. Note that in this study, we consider only a generic case in which we can assume a normal distribution of unknown mean $\mu$ and standard deviation $\sigma$. In the original BOCPD scheme, a reasonable next attempt to cope with this situation might be to take the derivative of the original time series and cast the problem as one of "change point detection of change rates" [1]. However, although it may be useful to use the derivative to detect change points, some information may be lost in this way compared to the original time series. This is because the original time series and a differentiated time series will interact differently with the underlying distributional assumptions when forming the run-length probability distribution. To fully utilize the information from the original time series, it would be ideal to be able to evaluate change points based on the original time series as well as a derived time series, such as a differentiated or integrated series, with minimal restrictions.

To elaborate the problem, two distinct data sets are used, as summarized in Table 1. The corresponding change point detection results of the BOCPD algorithm are illustrated in Fig. 5. For both cases, each datum in each predetermined partition $\rho$ is sampled from a normal distribution with a variable mean $\mu$ and a common standard deviation $\sigma = 1$. The former case is intended to represent a time series without long-term baseline shifts (zero-centered fluctuation), while the latter represents a time series with long-term baseline shifts. Change points are established at constant intervals of $t = 10$ for both cases. Each subplot in Fig. 4(a) illustrates a case in which the time series tends to regress back to a constant baseline value (zero-centered fluctuation), and in this case, the change points are fully captured without delay by the original BOCPD algorithm. In contrast, the plots in Fig. 4(b) illustrate a case in which the baseline of the time series is constantly increasing. As the baseline shifts farther from its original level, the change points are more poorly captured. The primary cause of this is that the contrast of the predictive distribution of $x_t$ given the previous run length of $r_t = 1$, namely, $P(x_t|r_{t-1}, x_t^{(\rho)})$, is degraded when the baseline is located farther from the original baseline. This can be observed in the second row (2) of Fig. 4. Here, we have used the predictive distributions of $x_t$ with previous run lengths $r_{t-1} \in 1, 2, ..., 10$ to accurately understand the interaction between a specific run length $r_{t-1}$ and an incoming data point $x_t$, which defines the probability of whether the run length $r_t$ will continue to grow in the current time step. The farther the data points are located from the original baseline, the blurrier the color pattern of the $P(x_t|r_{t-1}, x_t^{(\rho)})$ distribution becomes, indicating lower contrast at the decision boundaries. Furthermore, this loss of contrast is articulated in Fig. 5 by comparing the shapes of the posterior distributions with and without baseline shifts. As revealed by the graphs in Fig. 5, the shapes of the predictive probabilities $P(x_t|r_{t-1} = 10, x_t^{(\rho)})$ at the change points are flatter in the case with baseline shifts, especially at $t = 81$. By definition, the mechanism of change point detection relies heavily on the contrast of $P(x_t|r_{t-1} = 1, x_t^{(\rho)})$ given the most recent data point $x_t$. In other words, the more abruptly the value of $P(x_t|r_{t-1} = 1, x_t^{(\rho)})$ changes with the next data point observation, the easier it is for the BOCPD algorithm to detect a change point.

**Table 1. Data sets with and without baseline shifts.** Data sets 1 (without baseline shifts) and 2 (with baseline shifts) each consist of 100 data points. Each data point is generated from a normal distribution, with different means $\mu$ and the same standard deviation $\sigma = 1$ in the ten different partitions $\rho$. Change points are established in 9 locations, separated by a constant interval of $t = 10$.

| Data set | $\mu$ | $\sigma$ | Change points $t$ |
|---|---|---|---|
| 1 | 0,10,0,-20,0,20,0,-30,0,30 | 1 | 11,21, ...,91 |
| 2 | 0,10,20,30,40,50,60,70,80,70 | | |

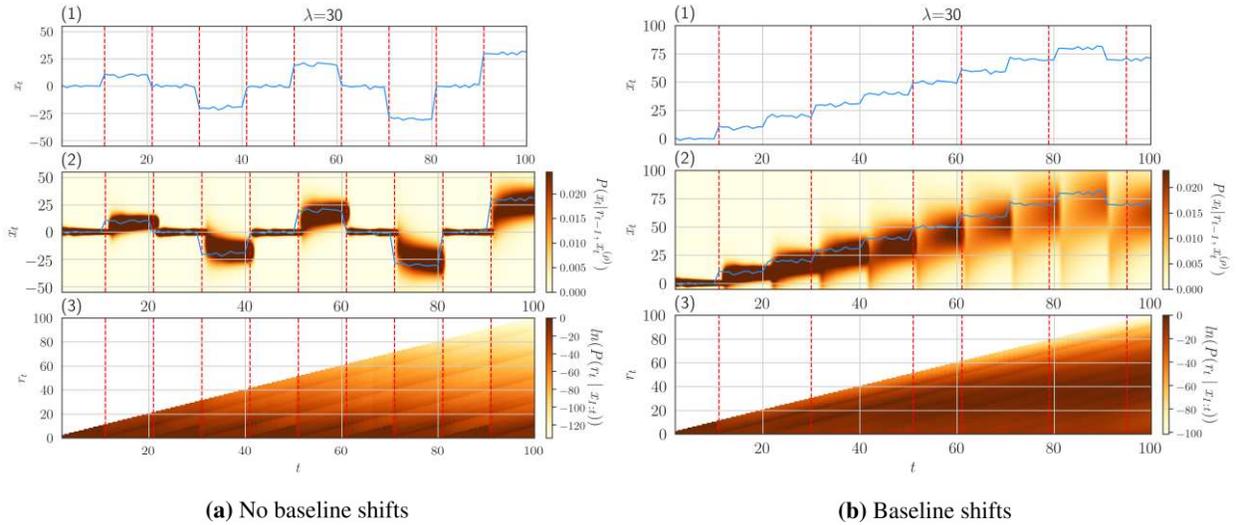

(a) No baseline shifts  (b) Baseline shifts

**Fig. 4. Raw data, predictive distributions, and run-length distributions with and without baseline shifts using the BOCPD algorithm.** Two different time series are used to demonstrate that the original BOCPD algorithm struggles to detect change points when there are baseline shifts. The plots in (a) illustrate a case of zero-centered fluctuation (without baseline shifts), in which the values cyclically regress to the baseline value of zero. The plots in (b) illustrate a case with baseline shifts, in which the values do not regress to the original baseline. The three plots on each side illustrate the following: (1) the raw data points; (2) the raw data points with the posterior predictive distribution of the observed data, $P(x_t|r_{t-1}, x_t^{(\rho)})$ (where $r_{t-1} \in \{1, 2, ..., 10\}$ is selected cyclically) in the background; and (3) the run-length distribution on a natural logarithmic scale, $ln(P(r_t \mid x_{1:t}))$. The vertical dotted lines in each plot indicate the detected change point locations.

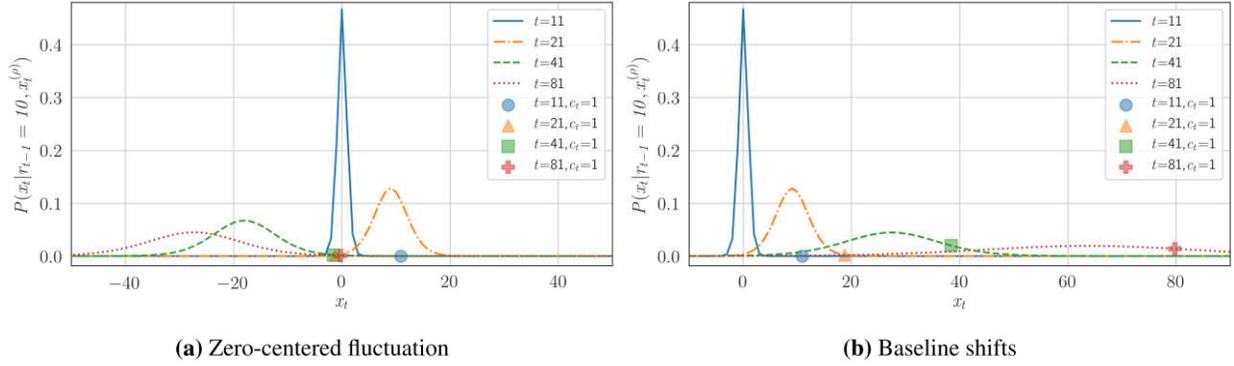

(a) Zero-centered fluctuation

(b) Baseline shifts

**Fig. 5. Shapes of the predictive distributions at times corresponding to change points.** The predictive probability values at the change points $c_t = 1$ at $t = \{11, 21, 41, 81\}$ are plotted on the predictive probability distributions $P(x_t | r_{t-1} = 10, x_t^{(\rho)})$ for a run length of $r_{t-1} = 10$ for the two cases presented in Fig. 4 ($\lambda = 30$ for both cases).

## 4 Solution

### 4.1 BOCPD for Baseline Shifts (BOCPD-BLS)

Our primary interest in this work lies in cases in which the baseline irreversibly shifts to a new level, such that an approximately long-term constant baseline cannot be assumed [7, 1, 5, 9, 11, 4]. To cope with data sets with baseline shifts, unlike most previous works, we regard all partitions $\rho$ as independent of each other. In other words, all parametric information from prior to the previous change point is discarded, without changing the underlying distributional nature of the observed data points. We call the modified algorithm proposed in the present work "BOCPD for baseline shifts (BOCPD-BLS)".

### 4.2 Parameter Initialization After Each Change Point

Once a change point is observed at $t(\in \mathbb{N})$, the following two steps are taken to feed the result ($c_t = 1$) back to a new detection process: 1) the parameters of the posterior predictive distribution, $\mu_t, \nu_t, \alpha_t$ and $\beta_t$ are initialized, and 2) the first observation in each partition $x_{ini}^{(\rho)}$, is reset to a value equal to the next observation $x_{t+1}$. Here, $x_{ini}^{(\rho)}$ is introduced to define the next partition's initial local baseline. Because we discard all the information from the previous partition except the knowledge of the existence of a change point, the run-length distribution and the joint distribution over the run length are rewritten as follows:

$$P(r_t \mid x_t^{'(\rho)}) = \frac{P(r_t, x_t^{'(\rho)})}{\sum_{r_t} P(r_t, x_t^{'(\rho)})} \tag{12}$$

and

$$P(r_t, x_r^{'(\rho)}) = \sum_{r_{t-1}} P(r_t \mid r_{t-1}) P(x_t' \mid r_{t-1}, x_t^{'(\rho)}) P(r_{t-1}, x_{t-1}') \tag{13}$$

where

$$x_{ini}^{(\rho=1)} = x_{t=1} \tag{14}$$

$$x_t' = x_t - x_{ini}^{(\rho)} \tag{15}$$

$$x_{t-1}' = \begin{cases} x_{t-1} - x_{ini}^{(\rho)} & (c_t = 0) \\ x_{t-1} - x_{ini}^{(\rho-1)} & (c_t = 1) \end{cases} \tag{16}$$

$$x_t'^{(\rho)} = \{x_{t-\tau}', x_{t-\tau+1}', ..., x_t'\} \qquad (\tau \in \mathbb{N}) \tag{17}$$

$$x_{1:t}' = \{x_1', x_2', ..., x_t'\} \tag{18}$$

$$\bigcup_\rho x_t'^{(\rho)} = x_{1:t}' \tag{19}$$

When making a new decision about the existence of a change point, we marginalize all past decisions $c_{1:(t-1)}$ out in the calculation of $P(r_t \mid x_t'^{(\rho)})$. Instead, $c_t$ is used purely to decide whether $x_{ini}^{(\rho)}$ is replaced by a new value. An example of a possible initial set of parameter values is $\mu_1 = x_1'$ and $\nu_1 = \alpha_1 = \beta_1 = 1$, and this set of initial parameters is utilized throughout this study. Since we intend to incorporate the result of change point detection after each observation into the beginning of the next partition, $\delta$ may now be expressed as

$$\delta = argmax(P(r_t \mid x_t'^{(\rho)})) - argmax(P(r_{t-1} \mid x_{t-1}'^{(\rho)})) \qquad (-(r_t - 1) \leq \delta \leq 1) \tag{20}$$

This indicates that the run-length distribution from the previous partition, $P(r_t \mid x_t'^{(\rho-1)})$, no longer affects the decision-making process regarding the existence of a change point in the current partition. As shown in Fig. 6, BOCPD-BLS uses the change point information $c_t$ to determine whether the parameters $\eta_\rho$ will be reinitialized (if $\delta \leq 0$, then $c_t = 1$) or continuously updated (if $\delta = 1$, then $c_t = 0$) before a newly incoming data point is observed. This additional process is the primary difference between BOCPD-BLS and the original BOCPD algorithm. The detailed calculation steps are described in Algorithm 1. The computational cost of BOCPD-BLS is equivalent to that of the original BOCPD algorithm, which is linear in the data count.

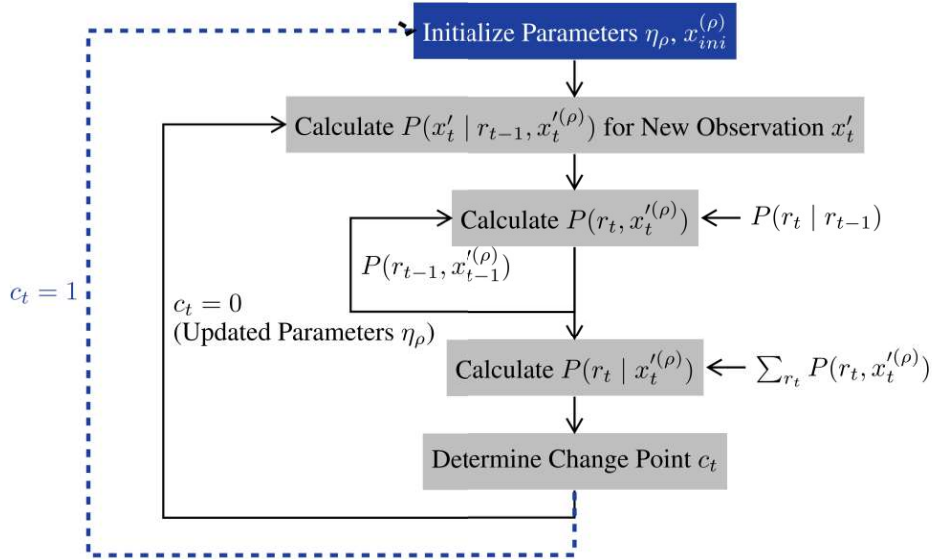

**Fig. 6. The BOCPD-BLS algorithm.** The white text and dashed line represent the major differences with respect to the original BOCPD algorithm. In the BOCPD-BLS algorithm, all parameters for calculating the predictive probability $P(x_t' \mid r_{t-1}, x_t'^{(\rho)})$ are reinitialized when a change point is observed such that the new partition $\rho$ is able to start from a new baseline.

## 5 Analysis

We prepared three data sets to assess the characteristics of the BOCPD-BLS algorithm compared to the original BOCPD algorithm in the presence of irreversible baseline shifts. The first data set is the same synthetic data set with multiple upward baseline shifts utilized in previous sections. The second tracks the history of the Bitcoin/USD exchange price, which is logged every minute. The third example is the number of confirmed positive COVID-19 cases in Russia from Jan. 22, 2020, through Oct. 6, 2020, on a daily basis.

### 5.1 Baseline Shifts: Synthetic Data

The generative parameters for this synthetic data set are listed in (data set 2 in this table). As seen from Fig. 7, BOCPD-BLS clearly captures all the change points without delay, while the original BOCPD algorithm shows some delay or misses the change point entirely for more than half of the predetermined change points. The primary reason for this difference is illustrated in Fig. 8, which highlights the difference in contrast degradation, especially when $x_t$ is far from the original baseline. By reinitializing the parameters of the predictive distribution after each change point, BOCPD-BLS is able to detect all change points without delay regardless of the $x_t$ values.

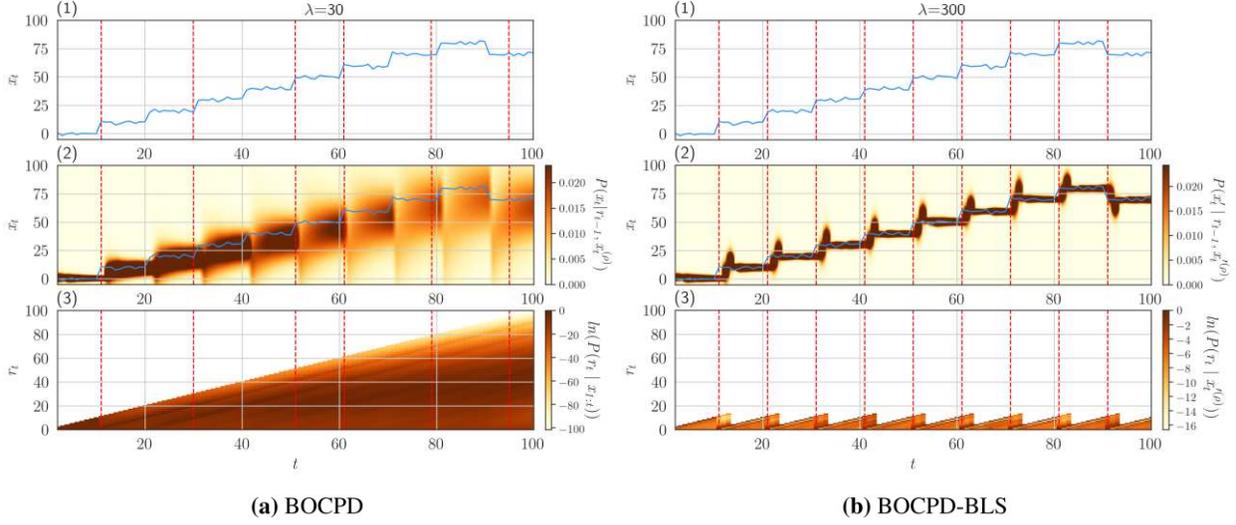

(a) BOCPD           (b) BOCPD-BLS

**Fig. 7. Change point detection using synthetic data set 2 in Table 1.** The $\lambda$ values in both cases are approximately optimal.

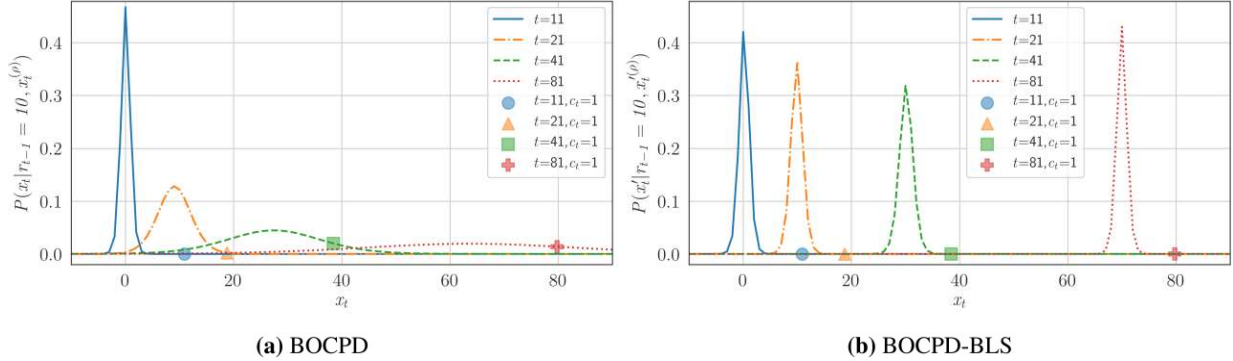

(a) BOCPD           (b) BOCPD-BLS

**Fig. 8. Comparison of the shapes of the predictive distributions at the change points.** (a) The contrast of the distribution is significantly degraded when $x_t$ is located farther from the original baseline ($t = 81$). (b) Each distribution maintains high contrast regardless of the $x_t$ value since the baseline of the underlying distribution is constantly updated after each change point.

### 5.2 Bitcoin vs. USD Price Data

As a real-world example, publicly available Bitcoin/USD price history data [15] are used. This data set displays upward baseline shifts over the long term. As shown in Fig. 9, the value starts near 9000 and exhibits multiple change points with continuous upward baseline shifts over the long run. Both major and minor mean shifts as well as slope shifts appear to exist. The plots in Fig. 9 show that BOCPD is able to detect the major upward mean shifts, while BOCPD-BLS captures not only the major mean shifts and most of the minor mean shifts but also several slope shifts, even at elevated $x_t$ levels. The $\lambda$ values used in both cases are the same as in the previous synthetic example presented in Fig. 8. For this reason, there may be some false positives or misses in both cases, but the main contrast information used in detecting minor mean and slope shifts with elevated $x_t > 10000$ appears to be valid. Moreover, it is important to note that the reason why the BOCPD algorithm misses the minor mean shifts is not because the magnitude of such a shift itself is small but rather because the values before and after the shift are both far from the original baseline, resulting in lower contrast of the run-length distribution $P(x_t|r_{t-1}=1, x_t^{(\rho)})$, as shown in Fig. 9(a)(2). We have checked the results with $\lambda$ value as low as 3, but the BOCPD algorithm cannot capture these minor shifts at elevated $x_t$ levels.

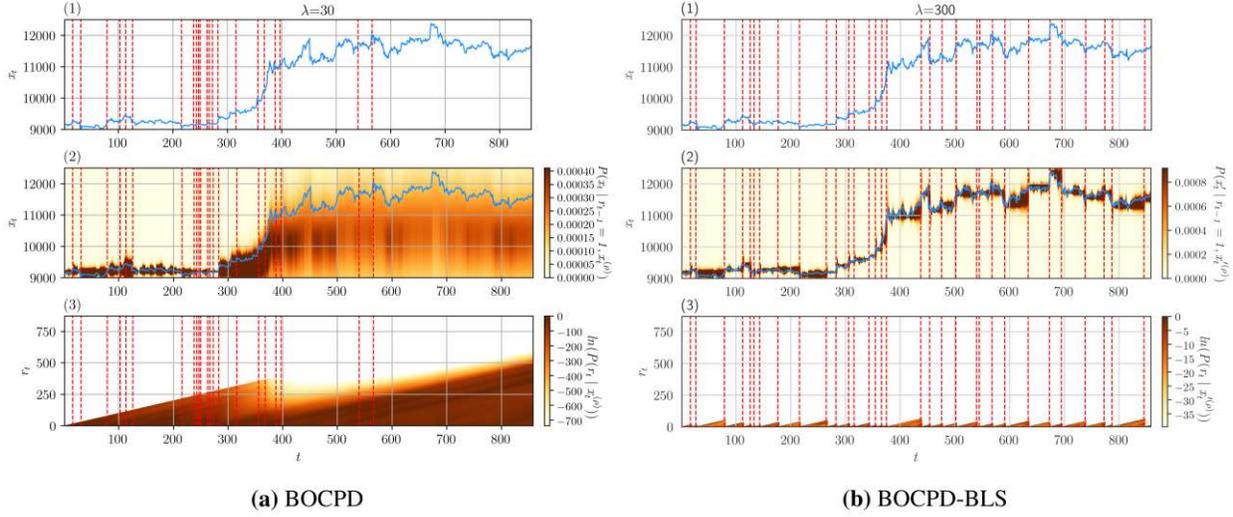

(a) BOCPD  (b) BOCPD-BLS

**Fig. 9. Change point detection using Bitcoin-USD price data (July 1, 2020 - August 31, 2020, data interval of 100 minutes.)** Although the BOCPD algorithm, as seen in (a), is able to capture the major mean shifts and some minor mean shifts in the first half of the data set, before the strongest elevation in $x_t$ occurs, it misses many minor mean shifts after this baseline shift. In contrast, the BOCPD-BLS algorithm, as seen in (b), detects not only the major mean shifts, but also most of the minor mean shiftss even after the strong elevation in $x_t$. In both cases, the $\lambda$ values are the same as those in Fig. 8.

## 5.3 Daily Confirmed COVID-19 Cases in Russia

As another real-world example, publicly available daily data on confirmed COVID-19 cases in Russia are used [16, 17]. This data set also displays upward baseline shifts over the long term. As seen in Fig. 10, the change points found using the BOCPD algorithm tend to be densely concentrated in certain regions where $x_t$ is either closer to the original baseline or abruptly increasing. In contrast, the BOCPD-BLS algorithm sparsely detects change points regardless of the observed $x_t$ value throughout the entire period. Another distinctive difference between the results of the two algorithms is the existence of locally clustered change points. Because the $\lambda$ settings are different between the two algorithms in Fig. 10, we have also tested the dependence of $\lambda$ by varying its value, as shown in Fig. 11. As revealed in Fig. 11, although varying $\lambda$ causes the number of detected change points to change in both cases, it appears to have little impact on the overall tendencies. This result may imply that the difference cannot be primarily attributed to the $\lambda$ settings but rather is due to differences in the algorithms themselves. Furthermore, since the BOCPD algorithm considers the entire run-length distribution from the beginning to the current $t$, some run-length probabilities carried over from previous partitions still interfere with the current decision-making. However, in the BOCPD-BLS algorithm, this does not occur since this algorithm discards the run-length distribution information from all previous partitions, thus simplifying the decision-making. This robustness to noise may be an additional advantage of BOCPD-BLS, as demonstrated through this example.

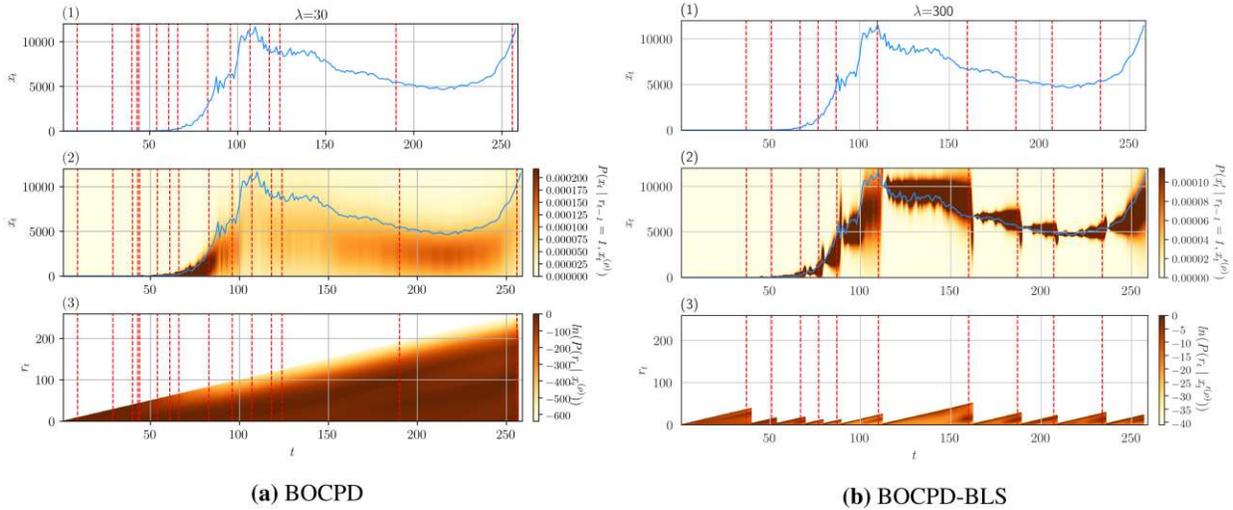

(a) BOCPD      (b) BOCPD-BLS

**Fig. 10. Change point detection using daily confirmed COVID-19 cases in Russia (January 22, 2020 - October 6, 2020).** (a) More change points tend to be detected when $x_t$ is relatively close to the original baseline or when $x_t$ sharply increases. (b) Change points are detected more sparsely throughout the entire period. In both cases, the $\lambda$ values are the same as those in Fig. 8.

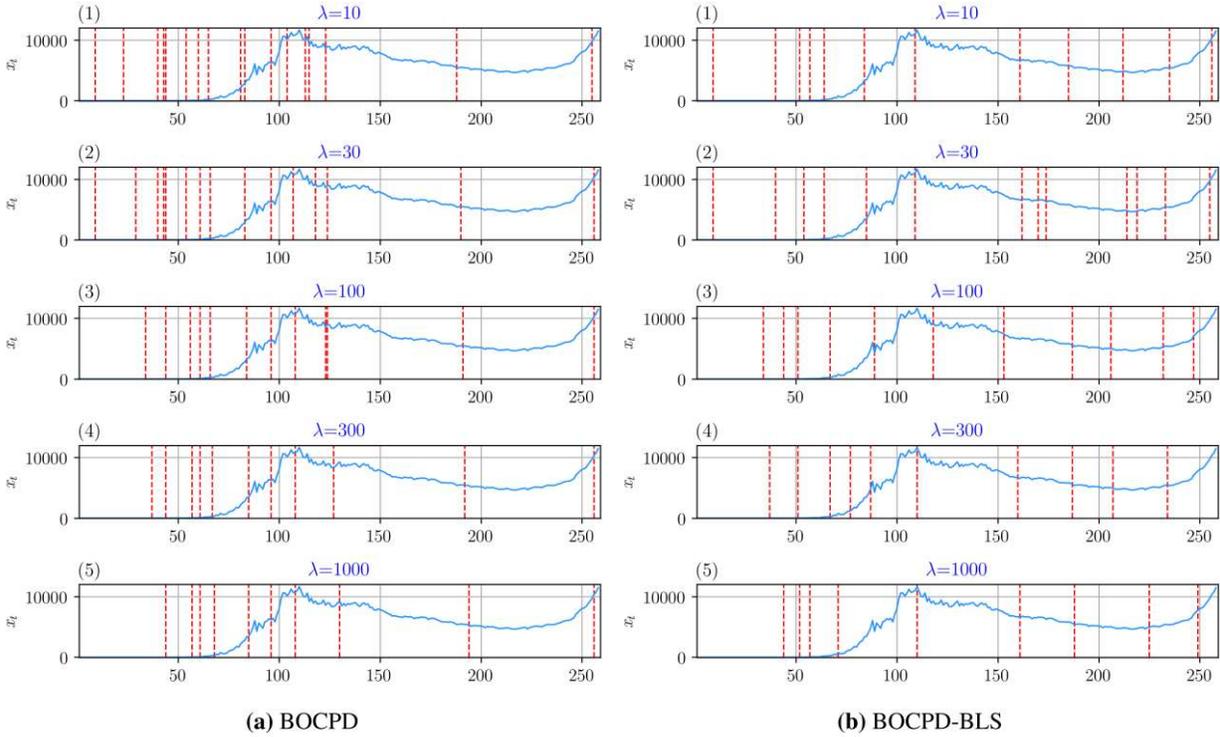

**Fig. 11. $\lambda$ dependence of change point locations.** The value of $\lambda$ is varied such that $\lambda \in \{10, 30, 100, 300, 1000\}$ to examine the overall balance of the change point locations using the same Russian COVID-19 data in Fig. 10. It is confirmed that (a) the BOCPD is only sensitive before the baseline shift occurs, while (b) the BOCPD-BLS algorithm's detection performance is balanced throughout the entire period.

## 6 Validation

For validation, we utilize six types of synthetic data sets with predetermined change points. These are designed primarily to assess the impacts of BOCPD-BLS compared with BOCPD in the presence of long-term baseline shifts; thus, data sets 2 and 6 are central to our interest (Table 2). However, the other data sets were additionally prepared to evaluate the differences in basic performance in the absence of long-term baseline shifts. Each data set was repeatedly generated with different pseudorandom seeds one hundred times from a normal distribution, and examples from each data set are displayed in Figure 12. To evaluate the change point detection capabilities of both algorithms, four metrics are adopted: 1) F-score (the F1 score, the harmonic mean of precision and recall), 2) Miss (the number of missed change points), 3) Delay (the amount of delay from the actual change point), and 4) Duplication (the number of duplicated change points detected in a partition, regardless of the delay in detection).

For simplicity, for data sets 1 through 4, a true positive (TP) is defined only as a change point for which the algorithm can detect the change point without a time delay, whereas for data sets 5 and 6, a time delay of up to $t = 5$ (within the first half of each partition) is allowed since detecting slope shifts appears to be more challenging. In other words, when an algorithm indicates change points outside of the allowed delay period, these are all regarded as false positive (FPs). Furthermore, if multiple change points are detected within the allowed delay period, this is regarded as a single TP detection for the F-score calculation on data sets 5 and 6 to avoid complexity.

As seen in Fig. 13, BOCPD-BLS demonstrates significantly better performance, particularly in the presence of the baseline shifts in data sets 2 and 6. Additionally, it shows equal or better performance for most combinations of metrics and data sets. It is important to note that although our primary focus is to assess whether BOCPD-BLS performs better on data sets 2 and 6 (both with baseline shifts), it also shows better performance and larger $\lambda$ margins in other cases without baseline shifts. Although the BOCPD-BLS algorithm appears to have the minor drawback that it tends to show a larger standard error of the mean, there are few cases in which BOCPD dominates overall. In summary, the following comments on each metric are offered based on Fig. 13. The mean statistics for $10 \leq \lambda \leq 1000$ are also summarized in Table 3 for quick reference.

**Table 2. Conditions of the six synthetic data sets used for performance validation.** Nine change points are consistently established at equal intervals of $t = 10$ at $t = \{11, 12, ...91\}$ for all data sets to form 10 partitions. Data set 1: Mean shifts centered around zero. Data set 2: Mean shifts with long-term baseline shifts. Data sets 3 and 4: First discrete differences for data sets 1 and 2, respectively. Data set 5: Slope shifts without baseline shifts. Data set 6: Slope shifts with baseline shifts. For data sets 5 and 6, we denote 0.1 by 0 and -0.1 by -0 for succinctness of expression.

| Data set | $\mu$ (or slope if noted) | $\sigma$ | Baseline shifts |
|---|---|---|---|
| 1 | 0,10,0,-20,0,20,0,-30,0,30 | 1 | No |
| 2 | 0,10,20,30,40,50,60,70,80,70 | 1 | Yes |
| 3 | First discrete differences of data set 1 | - | No |
| 4 | First discrete differences of data set 2 | - | No |
| 5 | (slope) 0,1,0,-1,0,2,0,-2,0,3 | 0.1 | No |
| 6 | (slope) -0,2,-0,2,-0,2,-0,2,-0,2 | 0.1 | Yes |

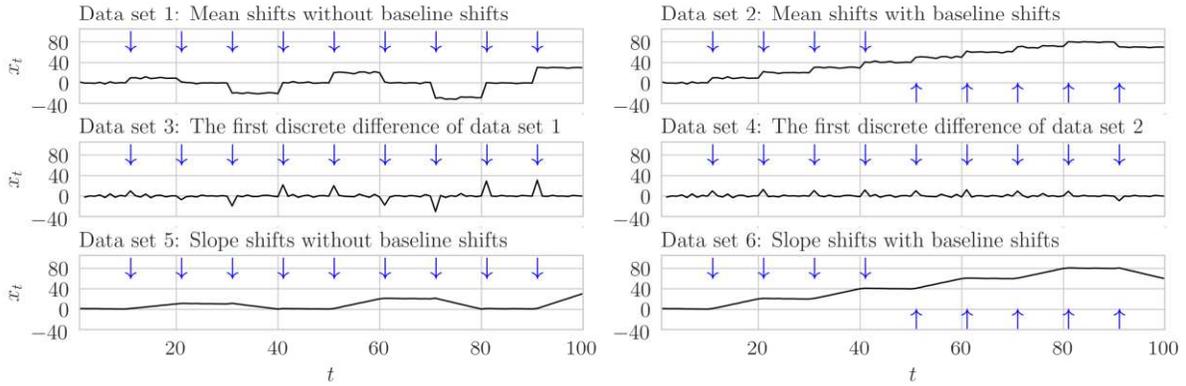

**Fig. 12. Six data sets used for validation tests.** The vertical arrows in each plot indicate the times of the predetermined change points.

- **F-score:** The BOCPD-BLS algorithm shows similar or higher scores for all data sets with various $\lambda$ values. Additionally, its performance degradation in the high $\lambda$ region is significantly smaller, especially on data set 1.

- **Miss:** The BOCPD-BLS algorithm shows fewer misses, particularly on the data sets with either baseline shifts or slope shifts (data sets 2, 5, and 6).

- **Delay:** The BOCPD-BLS algorithm shows less delay on the data sets with baseline mean shifts or the first discrete differences (data sets 2, 3, and 4).

- **Duplication:** No major difference is evident on all data sets, and the chance of duplicated detection is approximately 1 at most for all data sets and $\lambda$ values.

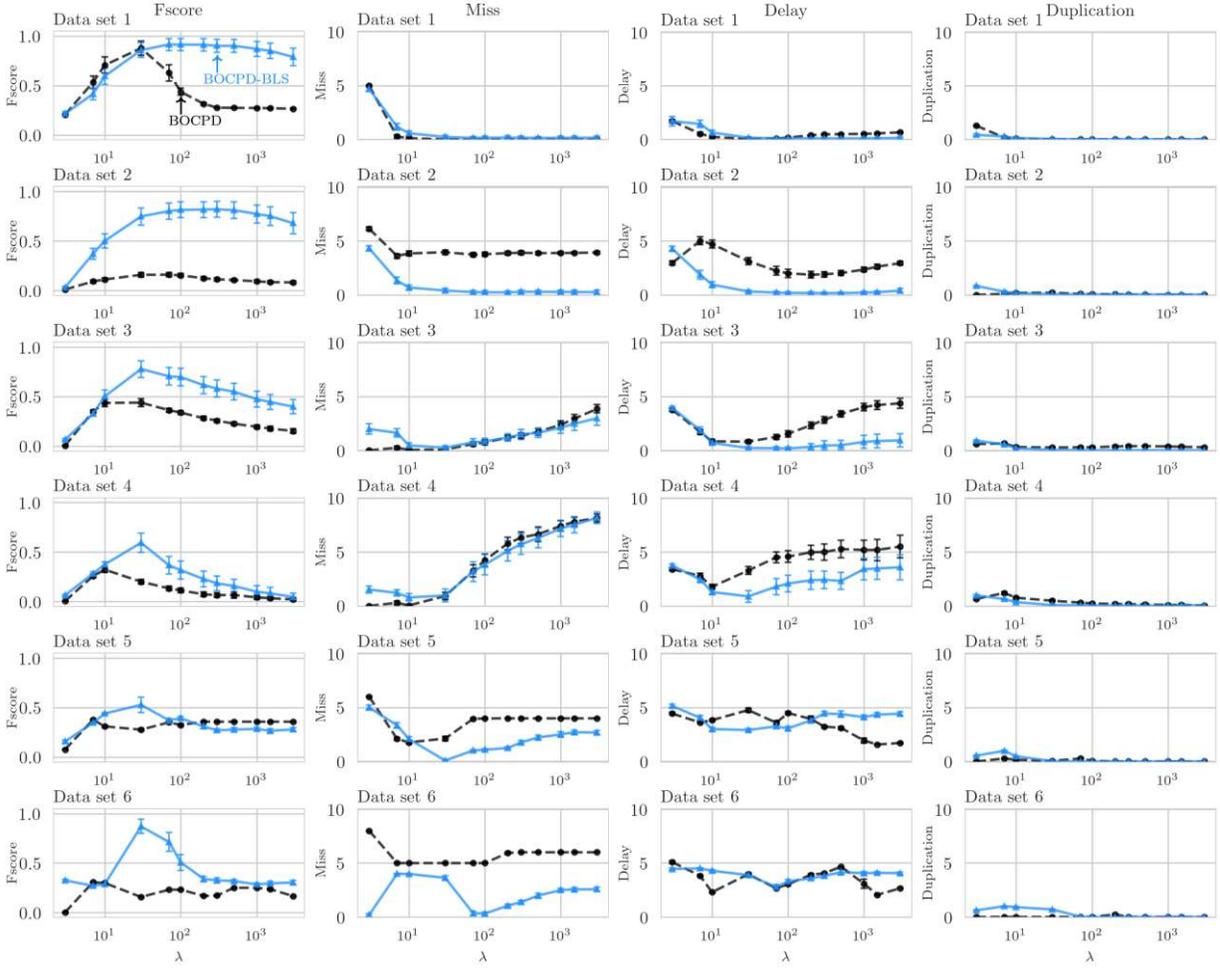

**Fig. 13. Fscore, Miss, Delay and Duplication on the six data sets.** Each data point consists of 100 sets of change point detection attempts on the synthesis data set with the different random seeds. An error bar in each data point indicates a standard error of the mean (SEM) multiplied by a factor of three ($3\sigma_{SEM}$).

| Algorithm | Data set | F-score | Miss | Delay | Duplication |
|---|---|---|---|---|---|
| BOCPD | 1 | 0.47 ±0.03 | **0.0** ±0.00 | 0.3 ±0.00 | **0.01** ±0.00 |
|  | 2 | 0.13 ±0.01 | 3.9 ±0.01 | 2.5 ±0.01 | 0.08 ±0.02 |
|  | 3 | 0.32 ±0.01 | **1.0** ±0.01 | 2.1 ±0.02 | 0.34 ±0.02 |
|  | 4 | 0.13 ±0.01 | 4.3 ±0.03 | 4.3 ±0.02 | 0.31 ±0.04 |
|  | 5 | 0.34 ±0.00 | 3.5 ±0.01 | 3.6 ±0.01 | 0.07 ±0.01 |
|  | 6 | 0.22 ±0.01 | 5.5 ±0.01 | **3.5** ±0.01 | **0.03** ±0.01 |
| BOCPD-BLS | 1 | **0.86** ±0.03 | 0.2 ±0.01 | **0.2** ±0.00 | 0.04 ±0.01 |
|  | 2 | **0.76** ±0.03 | **0.3** ±0.01 | **0.3** ±0.01 | **0.06** ±0.01 |
|  | 3 | **0.61** ±0.03 | 1.1 ±0.01 | **0.4** ±0.01 | **0.04** ±0.01 |
|  | 4 | **0.29** ±0.03 | **4.1** ±0.04 | **1.9** ±0.03 | **0.1** ±0.02 |
|  | 5 | **0.36** ±0.01 | **1.5** ±0.01 | 3.6 ±0.01 | 0.07 ±0.02 |
|  | 6 | **0.46** ±0.03 | **1.9** ±0.01 | 3.7 ±0.01 | 0.2 ±0.04 |

**Table 3. The mean values of the F-score, the number of missed change points, the delay, and the number of duplicated detections in each partition.** Each value represents 800 detection attempts (eight $\lambda$ values ($10 \leq \lambda \leq 1000$) and one hundred iterations with random seeds). The better performance in terms of each metric on each data set is highlighted in bold. The ± sign precedes the standard error of the mean (SEM) multiplied by a factor of three ($3\sigma_{SEM}$). The F-scores for all data sets show favorable results for the BOCPD-BLS algorithm.

# 7 Conclusions

The present work contributes to extending the application of BOCPD to time series data sets with mean and slope shifts in the presence of long-term baseline shifts toward arbitrary directions. The proposed extension of the BOCPD algorithm, called BOCPD-BLS, successfully adapts to such situations by feeding back information on a detected change point to guide the subsequent detection activities. This feedback process enables the algorithm to reinitialize the underlying baseline distribution, allowing it to maintain high detection sensitivity regardless of the offset of the $x_t$ values compared to the original baseline. Through a validation study, the proposed method has been confirmed to be particularly effective in the presence of baseline shifts, but it is also found to make the detection results less sensitive to the $\lambda$ value, thus enabling better performance even in the absence of baseline shifts.

# References

[1] Ryan Prescott Adams and David J. C. Mckay, Bayesian Online Changepoint Detection, arXiv:0710.3742v1 [stat.ML], 2007.

[2] Samaneh Aminikhanghahi and Diane J. Cook, A survey of methods for time series change point detection, Knowl. Inf. Syst. 51, 339-367, 2017.

[3] Amadou Ba and Sean A. McKenna, Water quality monitoring with online change-point detection methods, Journal of Hydroinformatics 17.1, 2015.

[4] Hon Fai Lau and Shigeru Yamamoto, Bayesian Online Changepoint Detection to Improve Transparency in Human-Machine Interaction Systems, 49th IEEE Conference on Decision and Control, 2010.

[5] Rakesh Malladi, Giridhar P Kalamangalam, Behnaam Aazhang, Online Bayesian Change Point Detection Algorithms for Segmentation of Epileptic Activity, IEEE Asilomar Conference on Signals, Systems and Computers, 2013.

[6] Alan H. Gee, Joshua Chang, Joydeep Ghosh, David Paydarfar, Bayesian Online Changeopint Detection of Physiological Transitions, In: Annual International Conference of the IEEE Engineering in Medicine and Biology Society (EMBC), 2018.

[7] Octavian Niculita, Zakwan Skaf, Ian K. Jennions, The application of Bayesian Change Point Detection in UAV Fuel Systems, 3rd International Conference on Through-life Engineering Services, 2014.

[8] Ryan Turner, Bayesian Change Point Detection For Satellite Fault Prediction, Proceedings of the interdisciplinary graduate conference, Cambridge University, UK, 2010.

[9] Ryan Turner, Yunus Saatci, Carl Edward Rasmussen, Adaptive Sequential Bayesian Change Point Detection, In Advances in Neural Information Processing Systems (NeurIPS): Temporal Segmentation Workshop, 2009.

[10] Eric Ruggieri and Marcus Antonellis, An exact approach to Bayesian sequential change point detection, Computational Statistics and Data Analysis 97 71-86, 2016.

[11] Diego Agudelo-Espana, Sebastian Gomez-Gonzalez, Stefan Bauer, Bernhard Scholkopf, Jan Peters, Bayesian Online Detection and Prediction of Change Points, arXiv:1902.04524v1 [cs.LG], 2019.

[12] Jean-Francois Ducre-Robitaille, Lucie A. Vincent, Gilles Boulet, Comparison of techniques for detection of discontinuities in temperature series, International Journal of Climatology, Int. J. Climatol. 23: 1087-1101, 2003.

[13] Naoki Shimada, Time Series Analysis, Kyoritsu Shuppan, ISBN 978-4-320-12501-8, 2019.

[14] K. P. Murphy, Conjugate Bayesian analysis of the gaussian distribution, Technical report, 2007.

[15] Zielak, Bitcoin Historical Data, Version 4, License: Creative Commons Attribution 4.0 International (CC BY-SA 4.0), Retrieved October 22, 2020 from
https://www.kaggle.com/mczielinski/bitcoin-historical-data

[16] COVID-19 Data Repository by the Center for Systems Science and Engineering (CSSE) at Johns Hopkins University (https://github.com/CSSEGISandData/COVID-19). License: Creative Commons Attribution 4.0 International (CC BY 4.0), Retrieved October 22, 2020 from
https://github.com/CSSEGISandData/COVID-19/blob/master/csse_covid_19_data/csse_covid_19_time_series/time_series_covid19_confirmed_global.csv

[17] Dong E, Du H, Gardner L. An interactive web-based dashboard to track COVID-19 in real time. Lancet Inf Dis. 20(5):533-534. doi: 10.1016/S1473-3099(20)30120-1.

# 8 Appendix: BOCPD-BLS Algorithm

The BOCPD-BLS algorithm is a simple extension of the original BOCPD algorithm. Algorithm 1 describes a batch version for detecting all change points using all data from the beginning of a time series until the most recent datum. For online detection, all parameters need to be cached for continuous reuse as each new datum is observed. As shown in Algorithm 1, the core idea of detecting change points based on the run-length distribution is inherited from the original BOCPD algorithm; the modifications focus on peripheral components that communicate the existence of a change point to the next partition such that the parameters of the underlying predictive distribution can be reinitialized to adapt to the new level of the baseline.

**Algorithm 1** BOCPD-BLS (* indicates a newly added component relative to the original BOCPD algorithm)

1: $\rho \Leftarrow 1$ {* The number of partitions}
2: $c_t \Leftarrow 1$ {* 1 if a change point is detected}
3: $cps \Leftarrow \emptyset$ {* An array for storing change point indices}
4: **for** $t = 1 : T$ **do**
5: $\quad$ Observe a new datum $x_t$
6: $\quad$ **if** $c_t = 1$ **then**
7: $\quad\quad$ $x_{ini}^{(\rho)} \Leftarrow x_t$ {* Store the first datum in a partition as the new baseline}
8: $\quad\quad$ $x'_t \Leftarrow x_t - x_{ini}^{(\rho)}$ {* Subtract the baseline from the observed datum}
9: $\quad\quad$ $\mu_t \Leftarrow x'_t$ {* Set the mean of the predictive distribution}
10: $\quad\quad$ $x_t'^{(\rho)} \Leftarrow \emptyset$ {* An array for storing the data in a partition}
11: $\quad\quad$ Initialize $\nu_t, \alpha_t,$ and $\beta_t$ {*}
12: $\quad\quad$ $r_t \Leftarrow 0$ {* Reset run-length count}
13: $\quad$ **else**
14: $\quad\quad$ $x'_t \Leftarrow x_t - x_{ini}^{(\rho)}$ {* Subtract the baseline from the observed datum}
15: $\quad$ **end if**
16: $\quad$ $r_t \Leftarrow r_t + 1$ {Increase the run-length count}
17: $\quad$ $x_t'^{(\rho)} \Leftarrow$ add $x'_t$ {Store the new datum in the array}
18: $\quad$ $P(x'_t \mid r_{t-1}, x_t'^{(\rho)}) \Leftarrow t_{2\alpha}(x'_t \mid \mu_t, \frac{\beta_t(\nu_t+1)}{\nu_t \alpha_t})$ {Calculate the predictive probability}
19: $\quad$ $P(r_t = 0, x_r'^{(\rho)}) \Leftarrow \frac{1}{\lambda} P(x'_t \mid r_{t-1}, x_t'^{(\rho)}) P(r_{t-1}, x'_{t-1})$ {Run-length probability for $r_t = 0$}
20: $\quad$ $P(r_t = r_{t-1} + 1, x_r'^{(\rho)}) \Leftarrow (1 - \frac{1}{\lambda}) P(x'_t \mid r_{t-1}, x_t'^{(\rho)}) P(r_{t-1}, x'_{t-1})$ {Run-length probability for an increase in $r_t$}
21: $\quad$ $P(r_t \mid x_t'^{(\rho)}) \Leftarrow \frac{P(r_t, x_t'^{(\rho)})}{\sum_{r_t} P(r_t, x_t'^{(\rho)})}$ {Normalize the run-length probability}
22: $\quad$ $c_t \Leftarrow 0$ {* Reset to 0}
23: $\quad$ $\delta \Leftarrow argmax(P(r_t \mid x_t'^{(\rho)})) - argmax(P(r_{t-1} \mid x_{t-1}'^{(\rho)}))$ {* Calculate delta to determine the existence of a change point}
24: $\quad$ **if** $\delta \leq 0$ **then**
25: $\quad\quad$ $c_t \Leftarrow 1$ {* Change point detected}
26: $\quad\quad$ $\rho \Leftarrow \rho + 1$ {* Update to a new partition}
27: $\quad\quad$ $cps \Leftarrow$ add $argmax(P(r_t \mid x_t'^{(\rho)}))$ {* Store the identified change point index}
28: $\quad$ **else**
29: $\quad\quad$ $\mu_{t+1} \Leftarrow \frac{\nu_t \mu_t + x}{\nu_t + 1}$ {Update the parameter}
30: $\quad\quad$ $\nu_{t+1} \Leftarrow \nu_t + 1$ {Update the parameter}
31: $\quad\quad$ $\alpha_{t+1} \Leftarrow \alpha_t + \frac{1}{2}$ {Update the parameter}
32: $\quad\quad$ $\beta_{t+1} \Leftarrow \beta_t + \frac{\nu_t(x - \mu_t)^2}{2(\nu_t + 1)}$ {Update the parameter}
33: $\quad$ **end if**
34: **end for**